%% file: main.tex
\documentclass[runningheads]{llncs}
\usepackage[T1]{fontenc}
\usepackage[dvipsnames,table]{xcolor}
\usepackage{colortbl}
\usepackage{graphicx}
\usepackage{booktabs}
\usepackage{import}
\usepackage{amssymb}
\usepackage{multirow}
\usepackage{pifont}
\usepackage{tabularx}
\usepackage{amsmath}
\usepackage{enumitem}
\usepackage[utf8]{inputenc}

\input{collections/modifications}

\begin{document}
\title{Mitigating representation bias caused by \newline missing pixels in methane plume detection}

\titlerunning{Mitigating bias caused by missing pixels in methane plume detection}

\author{Julia W\k{a}sala\inst{1,2}\orcidID{0000-0002-6352-8625} \and Joannes D. Maasakkers\inst{2}\orcidID{0000-0001-8118-0311} \and Ilse Aben\inst{2,3}\orcidID{0000-0003-2198-0768} \and Rochelle Schneider\inst{4}\orcidID{0000-0002-2905-0154} \and Holger Hoos\inst{5}\orcidID{0000-0003-0629-0099}\and Mitra Baratchi\inst{1}\orcidID{0000-0002-1279-9310}}
\authorrunning{J. W\k{a}sala et al.}
\institute{Leiden Institute for Advanced Computer Science (LIACS), Leiden, the Netherlands \and
SRON Space Research Organization Netherlands, Leiden, the Netherlands \and Department of Earth Sciences, Vrije Universiteit Amsterdam, Amsterdam, the Netherlands \and $\Phi$-lab, ESA-ESRIN, Frascati, Italy \and Chair for AI Methodology (AIM) at RWTH Aachen, Aachen, Germany. \\
\email{\{j.wasala\}@liacs.leidenuniv.nl}}
\textsc{Accepted as a workshop paper at MACLEAN - ECML/PKDD 2025.}

{\let\newpage\relax\maketitle}           % typeset the header of the contribution
\begin{abstract}

Most satellite images have systematically missing pixels (i.e., missing data not at random (MNAR)) due to factors such as clouds. 
If not addressed, these missing pixels can lead to representation bias in automated feature extraction models.
In this work, we show that spurious association between the label and the number of missing values in methane plume detection can cause the model to associate the coverage (i.e., the percentage of valid pixels in an image) with the label, subsequently under-detecting plumes in low-coverage images. 
We evaluate multiple imputation approaches to remove the dependence between the coverage and a label. 
Additionally, we propose a weighted resampling scheme during training that removes the association between the label and the coverage by enforcing class balance in each coverage bin. 
Our results show that both resampling and imputation can significantly reduce the representation bias without hurting balanced accuracy, precision, or recall.
Finally, we evaluate the capability of the debiased models using these techniques in an operational scenario and demonstrate that the debiased models have a higher chance of detecting plumes in low-coverage images. 

\end{abstract}
\section{Introduction}\label{sec:intro}
\import{sections/}{1_intro}

\section{Related Work}\label{sec:related}
\import{sections/}{2_related}

\section{Data}\label{sec:data}
\import{sections/}{3_data}

\section{Methods}\label{sec:methods}
\import{sections/}{4_methods}

\section{Empirical Evaluation Setup}\label{sec:empirical}
\import{sections/}{5_empirical}

\section{Results}\label{sec:results}
\import{sections/}{6_results}

\section{Conclusion}\label{sec:conclusion}
\import{sections/}{7_conclusion}

\begin{credits}
\subsubsection{\ackname} 
We thank Tobias de Jong, Berend Schuit, and Solomiia Kurchaba for their insightful feedback on our analyses and methodology. 
Parts of this research have been supported by: the “Physics-aware Spatio-temporal Machine Learning for Earth Observation Data” project (project number OCENW.KLEIN.425) of the Open Competition ENW research programme, which is partly financed by the Dutch Research Council (NWO);
the Open Space Innovation Platform\footnote{\url{https://ideas.esa.int}} as a Co-Sponsored Research Agreement and carried out under the Discovery programme of, and funded by, the European Space Agency (contract number 4000136204/21/NL/GLC/my); an Alexander von Humboldt Professorship in Artificial Intelligence awarded to Holger Hoos. Sentinel-5 Precursor is part of the EU Copernicus program, and Copernicus (modified) Sentinel-5P data (2018– 2023) have been used.

\end{credits}
\bibliographystyle{splncs04}
\bibliography{refs}

\end{document}

%% file: collections/modifications.tex
\newcommand{\hide}[1]{}

\definecolor{todocolor}{rgb}{0.9,0.1,0.1}
\definecolor{changedcolor}{rgb}{0.42,0.27,0.57}
\definecolor{juliacolor}{RGB}{16,38,148}
\definecolor{hhcolor}{RGB}{255,142,39}
\definecolor{mbcolor}{RGB}{17, 184, 184}
\definecolor{iacolor}{RGB}{251, 49, 153}
\definecolor{jdmcolor}{RGB}{135, 41, 150}
\definecolor{rscolor}{RGB}{119, 34, 6}

\newcommand{\cmark}{\ding{51}}%
\newcommand{\xmark}{\ding{55}}%

%% file: sections/1_intro.tex
Detecting and reducing large methane emissions is a promising approach towards mitigating global warming \cite{ipccreport,lauvaux_2022}. 
The TROPOMI satellite methane data product with daily global observations of atmospheric methane concentrations is a powerful resource for monitoring large emissions by detecting the associated methane plumes.
Training ML models for methane plume detection is challenging, because the TROPOMI methane data product contains missing pixels not at random (MNAR) over clouds or water, as methane concentrations can only be retrieved over water under specific circumstances \cite{lorente_2022}.
Since it is inevitable to have images with low coverage, it is important to be able to train models to detect plumes in scenes with missing pixels, such as over coastal areas. 

A simple approach to address missing pixels in satellite images is to remove images below a minimum coverage threshold before training. For higher resolution satellite images, such as Sentinel-2, missing pixels (e.g., due to cloud cover) can be imputed through interpolation or by using pixel values from a past scene \cite{arp_2024}. These solutions are, however, insufficient for methane plume detection in automated feature extraction-based models for the following reasons.
Firstly, many important regions where methane emissions occur have low coverage (more than $50\%$ of pixels missing), for example, because of proximity to coastlines.
Secondly, imputation algorithms based on static Earth surface properties are poorly suited to dynamic atmospheric problems, where emissions change and plumes move with the wind. 
Thirdly, because of the comparatively low spatial resolution of TROPOMI compared with Earth imaging satellites such as Sentinel-2 (i.e., $7\times5.5~\mbox{km}^2$ vs. $10\times10~\mbox{m}^2$), interpolation is likely to yield results comparable to single-value imputation. %\julia{no citation, this is from my own experience}
Finally, mindlessly imputing missing values can exacerbate representation bias. 

In this work, we demonstrate that failing to adequately handle missing pixels, particularly in combination with automated feature extraction, can cause the model to associate image coverage with methane plume presence, leading to under-detection in low-coverage images. This incorrect association is related to the concept of shortcut learning and confounders \cite{brown_2023}, a core concern in fair ML \cite{caton_2024}. 
We propose to address the representation bias due to MNAR by using data-centric approaches and present the following contributions:

\begin{itemize}
    \item We evaluate two deterministic imputation approaches and propose two new non-deterministic ones. We show that the choice of imputation strategy can impact representation bias. 
    \item We propose a resampling scheme during training to remove the dependence between coverage and the class label and show that this approach significantly improves bias-related metrics without hurting accuracy. We further show that combining resampling with imputation strategies leads to the strongest bias reduction.
    \item We evaluate the generalisation of the de-biasing approaches in an operational scenario and find that de-biased models increase the chance of finding plumes in low-coverage images.
\end{itemize} 

%% file: sections/2_related.tex
\textbf{Methane plume detection}. Schuit et al. \cite{schuit_2023} propose to detect methane plumes in TROPOMI methane data 
\cite{ATBDrpro} with a two-step ML pipeline that combines a CNN with an SVM, which reduces false positives by using physics-based features. 
This domain-informed approach shows robustness to image coverage, but lacks generalisability to related plume detection problems due to the methane-specific design.
W\k{a}sala et al. \cite{wasala_2025} enable the extension of this work to other gases by automatically designing end-to-end pipelines with neural architecture search.
ML approaches have also been proposed to detect plumes from individual facilities in (limited coverage) high-resolution satellite data ($\sim20~\mbox{m}$), such as from Sentinel-2
\cite{vaughan_2024a} and PRISMA \cite{joyce_2023}.
In this work, we build on the work by W\k{a}sala et al. \cite{wasala_2025} by designing new data-driven strategies inspired by approaches taken to address representation bias in Fair ML literature while maintaining the generalisability of automated feature extraction using neural networks. 

\textbf{Missing data and Fair ML}. Fernando et al. \cite{fernando_2021} show that simply removing instances with non-randomly missing features can exacerbate representation bias.
To address this bias, fair ML approaches propose different ways to deal with missing data.
Wang et al.~\cite{wang_2021c} propose an algorithm for weighted resampling of the dataset to account for non-random missingness in multi-class classification of tabular data.
Caton et al.~\cite{caton_2022} evaluate multiple imputation strategies and find that the choice of imputation can significantly impact fairness in tabular data classification.
While most approaches focus on tabular data, we address the issue of MNAR in satellite data. We explicitly treat the missing data as a confounder and propose a simple approach for addressing bias in model predictions by removing the dependence between coverage and the image label. % \mb{From the text it is not clear what is difference of your work with that of Wang et al.}.
% \mb{Aren't previous work also treating missing data as founder?}\julia{not explicitly, I think. I can weaken this statement}

%% file: sections/3_data.tex
\begin{figure}[t]
    \centering
    \includegraphics[width=\linewidth]{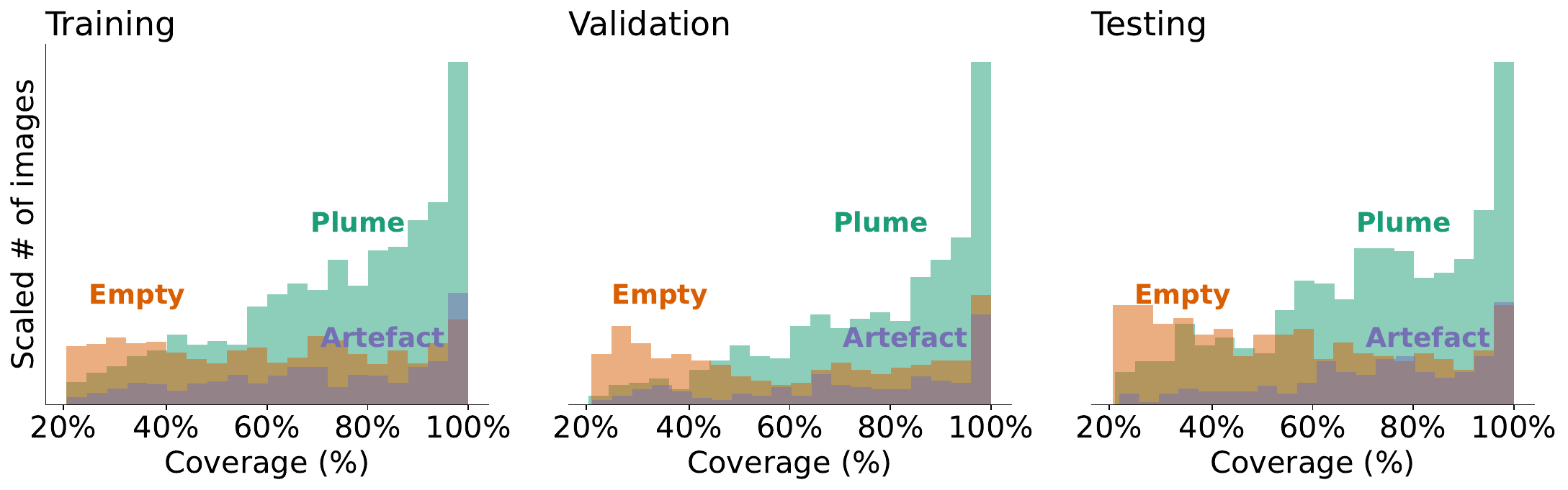}
    \caption{The distribution of images in the training, validation and testing partitions as a function of coverage (percentage of valid pixels per image). The partitions contain significantly more images of plumes with high coverage than low coverage. The y-axes are scaled independently across subplots to enable visual comparison.}
    \label{fig:train_dist}
\end{figure}

We detect methane plumes in a dataset of $9046$ images, each consisting of $32\times32$ pixel TROPOMI observations (Data version 19 \cite{ATBDrpro}).
The dataset extends the methane plume detection dataset created by Schuit et al. \cite{schuit_2023} with $~5000$ images detected using their model and verified by domain experts, creating a binary classification task ($56\%$ ``plume'' and $44\%$ ``not plume.'') \cite{schuit_2023a,dogniaux_2024} \footnote{\url{http://earth.sron.nl/methane-emissions}. Last accessed $19$ June $2025$. The public dataset only includes confirmed methane plumes, while we also use detections labelled as ``not plume.'' Data is available upon request.}.
The ``not plume'' class contains clearly empty images and artefacts, instances with plume-like features in the primary channel that are due to correlations with other retrieval parameters, such as albedo, that cause false positives.
To correctly classify these artefacts, we include six auxiliary data fields from the Sentinel-5P methane data product \cite{ATBDrpro}: surface pressure, albedo (SWIR), aerosol optical thickness (SWIR), data quality assurance values, cloud fraction (from Schuit et al. \cite{schuit_2023}), $\chi^2$ of the methane retrieval and land surface classification. We use spatial blocking ($3^\circ\times3^\circ$) to partition the data into $64\%$ training, $16\%$ validation, and $20\%$ testing data, preventing spatial leakage \cite{roberts_2017}.
Normalisation of the methane channel follows Schuit et al. \cite{schuit_2023}. The auxiliary channels are standardised using the training set means and standard deviations.
Most images in the dataset contain missing pixels, which introduce representation bias (Figure \ref{fig:train_dist}).  
We describe this bias and our mitigation approaches in the next section. 

%% file: sections/4_methods.tex
We propose two complementary approaches to reduce coverage bias to improve the detection of methane plumes in low-coverage images and make the code available.\footnote{\url{https://github.com/JuliaWasala/maclean-fair-ml-for-missing-pixels}} 
The dataset contains significantly more high- than low-coverage images of plumes (Figure \ref{fig:train_dist}), while non-plumes are uniformly distributed.
This representation bias creates a spurious relationship between the number of missing pixels and the label of an image, which the model can use to classify images based on the coverage of an image rather than features truly indicating the presence of plumes.
The simplest strategy to address missing pixels is to impute them: either using sophisticated methods \cite{arp_2024} that create realistic imputation values indistinguishable from real pixels, or with obvious placeholders that signal missing data to the model. 
However, imputation risks introducing additional noise into the problem. 
Another approach, entirely independent of imputation, is to sample training batches in a way to reduce the association between coverage and the label. 
When combined, imputation and resampling can address the problem on two levels: on an image level by replacing the missing pixels, and on a dataset level by changing the distribution of the dataset with respect to coverage and label. 
In the following, we describe our imputation and resampling approaches. 

\textbf{Imputing missing values.} We assume that no association should exist between image coverage and label.
While higher-coverage images contain more plumes due to having more pixels, the overall rarity of plumes suggests minimal impact from this assumption.
To remove the coverage dependence, we impute missing values in each image channel using two standard approaches and two novel methods proposed by us.
The novel approaches introduce non-determinism by sampling new pixel values at each epoch, making missing value locations unpredictable rather than learnable patterns. 
The four imputation approaches are the following: 
\begin{itemize}
    \item \textbf{Zero-imputation}:  Imputes each missing pixel with zero. This is a reasonable choice for the methane concentration because it signals low importance, but may skew the channel distributions with non-zero means.
    \item \textbf{Median-imputation}: Imputes each missing pixel with the median value of the channel. This approach avoids skewing the distributions and preserves valid categorical values, but both zero- and median-imputation risk creating artificial flat features.
    \item \textbf{Noise-augmented imputation (ours)}: Imputes the value of a missing pixel in each channel by sampling from a Gaussian distribution $\mathcal{N}(M'',\sigma) $  with channel median $M''$ 
    as mean and channel standard deviation $\sigma$. The added noise prevents the creation of large flat features when many adjacent pixels are missing. 
    \item \textbf{Pixel-sample imputation (ours)}: Imputes the value of each missing pixel by sampling values from the valid pixels in the image uniformly at random without replacement. 
\end{itemize}

\textbf{Resampling training data.} Resampling training data distributions during training is a common strategy for addressing imbalances in the data, such as class imbalance \cite{ghosh_2024}.
We combine undersampling and oversampling to achieve class balance in coverage bins, removing the statistical dependence between coverage and labels. 
We maintain the total number of images per coverage bin to avoid severe oversampling of the low-coverage images, which can lead to overfitting \cite{ghosh_2024}. 
We partition the training data into twenty equal-width bins based on the coverage of the images and calculate weights of each sample taken from that bin for each class as $ w_{i}^y = \frac{1}{|B_{i}^y|}\cdot \frac{|Bi|}{\sum_{j=0}^{19}{|B_j|}}$, 
where $B_{i}^y$ denotes the set of images in the $i^{th}$ coverage bin for each class label $y \in\{0,1\}$. 
We use these weights to draw new samples for each bin from the training data at each epoch, where each image's selection probability is proportional to its weight, ensuring the underrepresented class within each coverage bin is sampled more frequently to achieve class balance. 
For instance, given a dataset of $100$ images where coverage bin $B_1$ contains $2$ positive and $8$ negative examples, the sampling weights would be $0.05$ for the positives and $0.0125$ for negatives, leading to a balanced resampled distribution of $\approx 5$ positives and $\approx 5$ negatives in bin $B_1$. %\mb{Can you add something like this: leading to a balanced resampled distribution of x postive and negative samples after y times being resampled?}\julia{ok I need to think about phrasing because it's nondeterministic}. 

% \mb{This part is not clear and open to interpretation. Can you clarify: how are the weigth used as factor to draw x number of samples for each bin.}.

%% file: sections/5_empirical.tex
In the following, we describe the models used and the setup of our empirical evaluation. We aim to answer the following questions:
\begin{enumerate}[label=\textbf{Q\arabic*}]
    \item Does the association between coverage and labels affect classification performance, and which debiasing technique is most effective?
    \item  Do models trained to be less biased to coverage generalise better to operational scenarios? 
\end{enumerate}

We evaluated the effectiveness of our methods on two multi-image fusion architectures: (i) a vanilla CNN with six layers and (ii) a multi-branch CNN with one input branch for each data source (for details, see W\k{a}sala et al. \cite{wasala_2025}), though our de-biasing approach is architecture-agnostic. 
We trained the model for $50$ epochs with a batch size of $64$, the AdamW optimiser \cite{loshchilov_2019}, an initial learning rate of $1\cdot 10^{-5}$ and the cosine learning rate scheduler \cite{loshchilov_2017}.
We trained and evaluated each configuration $5$ times with different random seeds and applied the Mann-Whitney U test (suitable for small numbers of runs) to evaluate statistical significance between the top two model configurations.
All experiments ran on Leiden University's GRACE computing cluster, which features $26$ homogeneous CPU nodes with $94$ GBs of memory and Intel Xeon E5-2683 v4 CPUs ($2.10$GHz), and $9$ homogeneous GPU nodes with dual $2$ NVIDIA GeForce GTX 1080Ti configurations. 
All nodes operate under CentOS-7. 

\textbf{Evaluation metrics}:  We measured the performance in terms of two groups of metrics for each research question. 
To address Q1, we used a fully labelled dataset and measured the precision, recall, and balanced accuracy on the testing set. Additionally, we calculated two components of equalised odds \cite{caton_2024}: the difference in false positive rate (FPR) and true positive rate (TPR, which is equal to the difference in false negative rate) between $high$- and $low$-coverage images, given by  $\Delta TPR = TPR_{low} - TPR_{high}$ and $\Delta FPR = FPR_{low} - FPR_{high}$, respectively. 
To address Q2 and evaluate the performance in an operational scenario when no labels are available, we counted the number of images flagged as plume by each model ($\hat{y}>0.5$) and calculated the statistical parity \cite{caton_2024}, given by $\mbox{parity}=\frac{PR_{high}}{PR_{low}}$, where $PR_{high}$ and $PR_{low}$ are the positive classification rates for high and low coverage images.

%% file: sections/6_results.tex
% We first present the results of our approaches to address the representation bias in our dataset and then assess generalisation to an operational scenario. 

\begin{table}[t]
    \caption{Results of different de-biasing strategies Imputation (Imput.) and resampling (R.) on the hold-out test set (\textbf{Left}) and use-case application (\textbf{Right}). 
    % Both resampling at training time (R) and improved imputation strategies (i.e., median, sample, noise) significantly improve the equalised odds as measured by $\Delta FPR$ and $\Delta TPR$ (closer to $0$ is better) and use case parity, without hurting the balanced accuracy (BAcc), precision, and recall, compared to the default strategy without resampling and with zero imputation. 
    Significantly worst imputation strategy (per model and resampling) shown in \textcolor{red}{red}. Best scores per architecture are \textbf{bolded}. Significantly better performance within each model architecture (vanilla (V) and M-branch (M)) and imputation strategy pair is \underline{underlined}. BAcc stands for balanced accuracy.}
    % \centering
    \hspace{-0.5cm}
     \scriptsize
\begin{tabular}{llllllll|ll}
% \begin{tabular}{lll@{\hskip 0.1in}l@{\hskip 0.1in}l@{\hskip 0.1in}l@{\hskip 0.1in}l@{\hskip 0.1in}l@{\hskip 0.1in}|l@{\hskip 0.1in} l}
      
\toprule
\multicolumn{3}{l}{\textit{Research question}} & \multicolumn{5}{c|}{\textbf{Q1}} & \multicolumn{2}{c}{\textbf{Q2}}\\
Model & Imput. & R. & BAcc & Precision & Recall & $\Delta FPR$ & $\Delta TPR$ & Parity &Flags\\
\midrule
\multirow[t]{12}{*}{V}   & \multirow[t]{2}{*}{Zero} & \xmark & $0.73 \pm 0.01$ & $\mathbf{0.78 \pm 0.01}$ & $0.70 \pm 0.03$ & \textcolor{red}{$-0.24 \pm 0.02$} & \textcolor{red}{$-0.32 \pm 0.02$} & \textcolor{red}{$7.59 \pm 0.52$} & $5054$\\
  & & \cmark & $0.73 \pm 0.01$ & $0.77 \pm 0.01$ & $0.72 \pm 0.02$ & \underline{\textcolor{red}{$-0.08 \pm 0.03$}} & \underline{\textcolor{red}{$-0.08 \pm 0.03$}} & \underline{\textcolor{red}{$2.88 \pm 0.26$}} & $5764$\\

\rowcolor{lightgray!50}\cellcolor{white} & \multirow[t]{2}{*}{Median} & \xmark & $0.73 \pm 0.01$ & $0.77 \pm 0.01$ & $0.71 \pm 0.03$ & $-0.07 \pm 0.01$ & $-0.12 \pm 0.02$ & $2.38 \pm 0.13$ & $6546$\\
\rowcolor{lightgray!50}\cellcolor{white} & & \cmark & $0.73 \pm 0.02$ & $0.77 \pm 0.01$ & $0.70 \pm 0.04$ & $\mathbf{~0.01 \pm 0.03}$ & \underline{$\mathbf{-0.01 \pm 0.03}$} & \underline{$1.81 \pm 0.09$} & $6280$\\

   & \multirow[t]{2}{*}{Sample} & \xmark & $\mathbf{0.74 \pm 0.01}$ & $\mathbf{0.78 \pm 0.02}$ & $0.72 \pm 0.04$ & $-0.12 \pm 0.05$ & $-0.06 \pm 0.02$ & $2.61 \pm 0.24$ & $5813$\\
  & & \cmark & $\mathbf{0.74 \pm 0.00}$ & $\mathbf{0.78 \pm 0.01}$ & $0.73 \pm 0.02$ & \underline{$-0.03 \pm 0.03$} & \underline{$-0.01 \pm 0.01$} & \underline{$1.71 \pm 0.15$} & $6334$ \\

\rowcolor{lightgray!50}\cellcolor{white} & \multirow[t]{2}{*}{Noise} & \xmark & $\mathbf{0.74 \pm 0.01}$ & $0.77 \pm 0.01$ & $\mathbf{0.75 \pm 0.03}$ & $-0.19 \pm 0.01$ & $-0.13 \pm 0.03$ & $3.07 \pm 0.24$ & $6385$ \\
\rowcolor{lightgray!50}\cellcolor{white}& & \cmark & $\mathbf{0.74 \pm 0.01}$ & $0.77 \pm 0.01$ & $\mathbf{0.75 \pm 0.02}$ & \underline{$-0.03 \pm 0.04$} & \underline{$-0.03 \pm 0.02$} & \underline{$\mathbf{1.66 \pm 0.13}$} & $7171$\\

\midrule
\multirow[t]{12}{*}{M}   & \multirow[t]{2}{*}{Zero} & \xmark & $0.83 \pm 0.01$ & $0.86 \pm 0.01$ & $0.83 \pm 0.03$ & \textcolor{red}{$-0.14 \pm 0.02$} & \textcolor{red}{$-0.14 \pm 0.02$} & \textcolor{red}{$8.16 \pm 1.31$}& $2925$\\
  
  & & \cmark & $\mathbf{0.84 \pm 0.01}$ & $0.86 \pm 0.01$ & $0.85 \pm 0.02$ & \underline{$-0.09 \pm 0.03$} & \underline{$-0.05 \pm 0.02$} & \underline{\textcolor{red}{$3.40 \pm 0.62$}} & $2662$\\

\rowcolor{lightgray!50}\cellcolor{white} & \multirow[t]{2}{*}{Median} & \xmark & $\mathbf{0.84 \pm 0.01}$ & $\mathbf{0.87 \pm 0.01}$ & $0.84 \pm 0.02$ & $-0.12 \pm 0.02$ & $-0.11 \pm 0.02$ & $3.75 \pm 0.44$ & $3707$\\
\rowcolor{lightgray!50}\cellcolor{white} & & \cmark & $\mathbf{0.84 \pm 0.00}$ & $\mathbf{0.87 \pm 0.01}$ & $0.84 \pm 0.02$ & \underline{$-0.07 \pm 0.02$} & \underline{$-0.05 \pm 0.02$} & \underline{$2.02 \pm 0.30$} & $3621$\\

  & \multirow[t]{2}{*}{Sample} & \xmark & $\mathbf{0.84 \pm 0.00}$ & $0.85 \pm 0.01$ & \textbf{$0.86 \pm 0.01$} & $-0.12 \pm 0.02$ & $-0.07 \pm 0.01$ & $2.78 \pm 0.14$ & $3416$\\
  & & \cmark & $0.83 \pm 0.01$ & $0.86 \pm 0.01$ & $0.82 \pm 0.03$ & \underline{$\mathbf{-0.06 \pm 0.03}$} & $\mathbf{-0.02 \pm 0.03}$ & $2.03 \pm 0.27$ & $2933$\\

\rowcolor{lightgray!50}\cellcolor{white} & \multirow[t]{2}{*}{Noise} & \xmark & $0.82 \pm 0.01$ & $0.84 \pm 0.01$ & $0.85 \pm 0.02$ & $-0.12 \pm 0.02$ & $-0.11 \pm 0.03$ & $2.51 \pm 0.28$ & $4383$\\
\rowcolor{lightgray!50}\cellcolor{white} & & \cmark & $0.83 \pm 0.01$ & $0.84 \pm 0.01$ & $0.85 \pm 0.04$ & \underline{$-0.08 \pm 0.02$} & \underline{$-0.06 \pm 0.02$} & \underline{$\mathbf{1.69 \pm 0.12}$} & $4141$\\

\bottomrule
\end{tabular}
    \label{tab:main}
\end{table}

\subsection{Resampling and imputation reduce bias without hurting accuracy}
We train and evaluate methane plume detection networks on all combinations of resampling and imputation strategies (Table \ref{tab:main}, left). 
% and parity
Median, pixel-sample, and noise-augmented imputation yield significantly better $\Delta FPR$ and $\Delta TPR$, with the exception of the $\Delta TPR$ of the multi-branch model trained with resampling and median imputation, which does not significantly differ from zero imputation. 
Non-determinism is, therefore, not strictly necessary, as there are no significant differences between the three imputation strategies. 
Furthermore, filling pixels with exact values present in the data (as median and pixel-sample do) is also not necessary, because noise-augmented imputation performs equally well as median and pixel-sample imputation.
These results show that the right choice of imputation strategy can significantly affect the bias, but the networks we evaluated show similar performance across most imputation strategies. 

Networks with resampling all have significantly better $\Delta FPR$ and $\Delta TPR$ compared to those without resampling, and this holds for each imputation technique, except for the multi-branch model with pixel-sample imputation. %\jdm{why?} not sure why... i think its just randomness
Resampling did not affect the balanced accuracy, precision or recall for any of the networks, showing that reduced bias does not have to come at the cost of performance. %\jdm{Do we have the results without any imputation / resampling though?}\julia{yes, without resampling are the rows with x-es in the table. without imputation is not possible because you cannot train with NaNs. I guess zero-filling is closest thing}
Most configurations show a slightly negative $\Delta FPR$ and $\Delta TPR$, which indicates that the false and true positive rates are slightly higher in the high-coverage images and there are thus still relatively more high-coverage images flagged as plume.
This is a desired behaviour, because fully equalised odds between low- and high-coverage images are unreasonable to expect, since the high-coverage have more pixels that could feature plumes.

% \julia{there was a subsection here about analysing artefacts but I dont' think that's going to fit in here, because the table is also quite big and I need to explain a lot}

\begin{figure}[t]
    \centering
    \includegraphics[width=\textwidth]{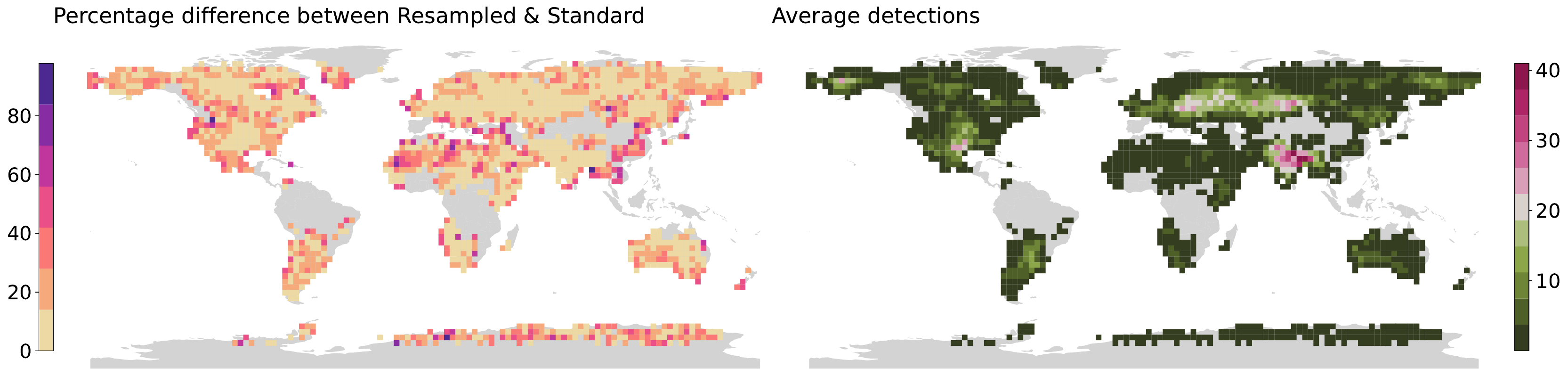}
    \caption{ (\textbf{Left}): Absolute percentage difference between average number of flags per grid cell (aggregated over architecture and imputation strategies) for resampled vs. standard (no resampling) networks. Dark purple cells indicate higher disagreement, though difference are small due to few average detections in those cells (\textbf{Right}, averaged over all models). Overall, the predictions show little difference between resampled and standard networks. Grey basemap/white background indicate no data or no difference. 
    }
    \label{fig:usecase}
\end{figure}

\subsection{Less biased models flag more low-coverage images as plume}
% \jdm{Furthermore, there are A LOT of detections in here that don’t make any sense south pole, middle of Australia where there’s nothing to emit. }
In operational scenarios, plumes should occur mostly independently of image coverage, unlike in our biased training distribution (Figure \ref{fig:train_dist}). % \mb{last part is unclear. what do you mean coverage-label distribution? do you mean the resampled distribution used for training}
% W\k{a}sala et al. \cite{wasala_2025} also found a large discrepancy between the precision on the testing partition and a use case analysis. 
Therefore, we apply each evaluated model to a previously unseen testing set, consisting of a week of TROPOMI methane data over land from $14-20$ March $2022$. 
We cropped $32\times32$ pixel images, with a shifted window with an offset of $16$ pixels and processed these following the procedure from Section \ref{sec:data}, yielding $20\,965$ images. 
We calculate the total number of images flagged as plume (thresholding the predicted scores at $0.5$), aggregated across network architecture and imputation strategies, to compare resampled versus standard (no resampling) models. 

% \mb{same comment about the figure here, just explain the visible facts from the figure}
We compare the difference between the average number of flags per $3\times3^\circ$ grid cell between resampled and standard training (averaged across networks and imputation strategies, Figure \ref{fig:usecase}).
The disagreement is small in most regions, and the average detection count (Figure \ref{fig:usecase}, right) in cells with high disagreement is low; therefore, the overall impact of these differences is small.
Both resampling and non-zero imputation significantly improve the parity (lower is better) of the use case detections (Table \ref{tab:main}, right), increasing the chance of finding plumes in low-coverage regions (such as coasts) that would otherwise be unfairly left out.
However, more research is needed to determine whether more plumes are actually detected in low-coverage images, as many of the flagged images may be false positives. The total number of images flagged is an order of magnitude higher than expected, compared to the validated 
detections\footnote{\url{http://earth.sron.nl/methane-emissions}. Last accessed $19$ June $2025$.} obtained with the model proposed by Schuit et al. \cite{schuit_2023}, and many of these flagged images occur in places where no known large methane emission sources exist (such as the South Pole). 
Saliency maps suggest our models have difficulty identifying relevant segments of and extracting features from the auxiliary channels, explaining why the approach by Schuit et al. \cite{schuit_2023} using domain-specific features is more robust to false positives than automated feature extraction approaches.

% Therefore, resampling and imputation increase the chance of detecting low-coverage images, which would otherwise be unfairly left out. \jdm{I’m not sure we can draw conclusions if there are this many false positives. It seems like the overwhelming majority of the detections would be a false positive.}

%% file: sections/7_conclusion.tex
In this work, we proposed two data-driven strategies to mitigate representation bias due to missing pixels in methane plume detection from TROPOMI satellite observations: (i) implementing multiple imputation strategies and (ii) resampling the training dataset during training to eliminate dependencies between coverage and image labels. 
Evaluation on a fully labelled test set showed that both approaches significantly improved equalised odds metrics without sacrificing balanced accuracy, precision or recall.
Simple imputation methods performed similarly to more complex alternatives. 
Evaluation in an operational scenario showed that while less biased models flagged more low-coverage images as plumes, the number of flagged images is higher than expected compared to a model using physics-based features, suggesting potential false positives caused by challenges in automated feature extraction from the auxiliary channels, which we aim to address in future work to improve the practical use of automated feature extraction models in methane plume detection.  